%% file: _main.tex
\input{_constants}
\arxiv 

\pdfoutput=1
\documentclass[10pt,twocolumn,letterpaper]{article}
\input{cvpr_header}

\unless\ifarxiv \myexternaldocument{_supplementary} \fi

\begin{document}
\title{HeAL3D: Heuristical-enhanced Active Learning for 3D Object Detection}
\author{Esteban Rivera\textsuperscript{*} \quad Surya Prabhakaran \quad Markus Lienkamp \\
Technical University of Munich and Munich Institute of Robotics and Machine Intelligence\\
Munich, Germany\\
{\tt\small name.lastname@tum.de}
}
\maketitle\let\thefootnote\relax\footnotetext{\textsuperscript{*}Corresponding author}
\input{00_abstract}

\input{01_intro}

\input{02_related}

\input{03_method}
\input{04_results}

\input{10_conclusion}
{\small
\bibliographystyle{ieeenat_fullname}
\bibliography{11_references}
}


\end{document}


\title{\paperTitle}
\author{\authorBlock}
\maketitlesupplementary

\appendix
\input{12_appendix}

{\small
\bibliographystyle{ieeenat_fullname}
\bibliography{11_references}
}

%% file: _constants.tex
\def\paperID{}
\def\confName{CVPR}
\def\confYear{2025}

\newif\ifreview 
\newif\ifarxiv \newcommand{\arxiv}{\arxivtrue}
\newif\ifcamera 
\newif\ifrebuttal 

%% file: cvpr_header.tex
\ifreview \usepackage[review]{cvpr} \fi
\ifarxiv \usepackage[pagenumbers]{cvpr} \fi
\ifrebuttal \usepackage[rebuttal]{cvpr} \fi
\ifcamera \usepackage{cvpr} \fi

\input{_macros}  

\usepackage{xr-hyper}

\makeatletter
\newcommand*{\addFileDependency}[1]{
  \typeout{(#1)}
  \@addtofilelist{#1}
  \IfFileExists{#1}{}{\typeout{No file #1.}}
}

\makeatother
\newcommand*{\myexternaldocument}[1]{
    \externaldocument{#1}
    \addFileDependency{#1.tex}
    \addFileDependency{#1.aux}
}

\definecolor{cvprblue}{rgb}{0.21,0.49,0.74}
\usepackage[pagebackref,breaklinks,colorlinks,allcolors=cvprblue]{hyperref}
\usepackage[capitalize]{cleveref}
\crefname{section}{Sec.}{Secs.}
\crefname{table}{Table}{Tables}
\crefname{figure}{Fig.}{Figs.}

\ifarxiv \crefname{appendix}{App.}{Apps.}
\else \crefname{appendix}{Suppl.}{Suppls.} \fi

%% file: _macros.tex
\usepackage{graphicx}	
\usepackage{amsmath}	
\usepackage{amssymb}	
\usepackage{booktabs}
\usepackage{times}
\usepackage{microtype}
\usepackage{epsfig}
\usepackage{caption}
\usepackage{float}
\usepackage{placeins}
\usepackage{color, colortbl}
\usepackage{stfloats}
\usepackage{enumitem}
\usepackage{tabularx}
\usepackage{xstring}
\usepackage{multirow}
\usepackage{xspace}
\usepackage{url}
\usepackage{subcaption}
\usepackage{xcolor}
\usepackage[hang,flushmargin]{footmisc}
\usepackage{amssymb}
\usepackage{array}
\ifcamera \usepackage[accsupp]{axessibility} \fi

% ...

\ifarxiv  \fi

\newcommand{\name}[1]{{\textcolor{blue}{[HeAL]}}}

\newcommand{\R}[1]{{%
    \textbf{%
        \ifstrequal{#1}{1}{\textcolor{red}{R#1}}{%
        \ifstrequal{#1}{2}{\textcolor{blue}{R#1}}{%
        \ifstrequal{#1}{3}{\textcolor{magenta}{R#1}}{%
        \ifstrequal{#1}{4}{\textcolor{teal}{R#1}}{%
                           \textcolor{cyan}{R#1}%
        }}}}%
    }%
}}

%% file: 00_abstract.tex
\begin{abstract}
Active Learning has proved to be a relevant approach to perform sample selection for training models for Autonomous Driving. Particularly, previous works on active learning for 3D object detection have shown that selection of samples in uncontrolled scenarios is challenging. Furthermore, current approaches focus exclusively on the theoretical aspects of the sample selection problem but neglect the practical insights that can be obtained from the extensive literature and application of 3D detection models.
In this paper, we introduce HeAL (Heuristical-enhanced Active Learning for 3D Object Detection) which integrates those heuristical features together with Localization and Classification to deliver the most contributing samples to the model's training.
In contrast to previous works, our approach integrates  heuristical features such as object distance and point-quantity to estimate the uncertainty, which enhance the usefulness of selected samples to train detection models. Our quantitative evaluation on KITTI shows that HeAL presents competitive mAP with respect to the State-of-the-Art, and achieves the same mAP as the full-supervised baseline with only 24\% of the samples.
\vspace{-0.5cm}
\end{abstract}

%% file: 01_intro.tex
\section{Introduction}
\label{sec:intro}
\begin{figure}[ht]
    \centering
    \includegraphics[width=\linewidth]{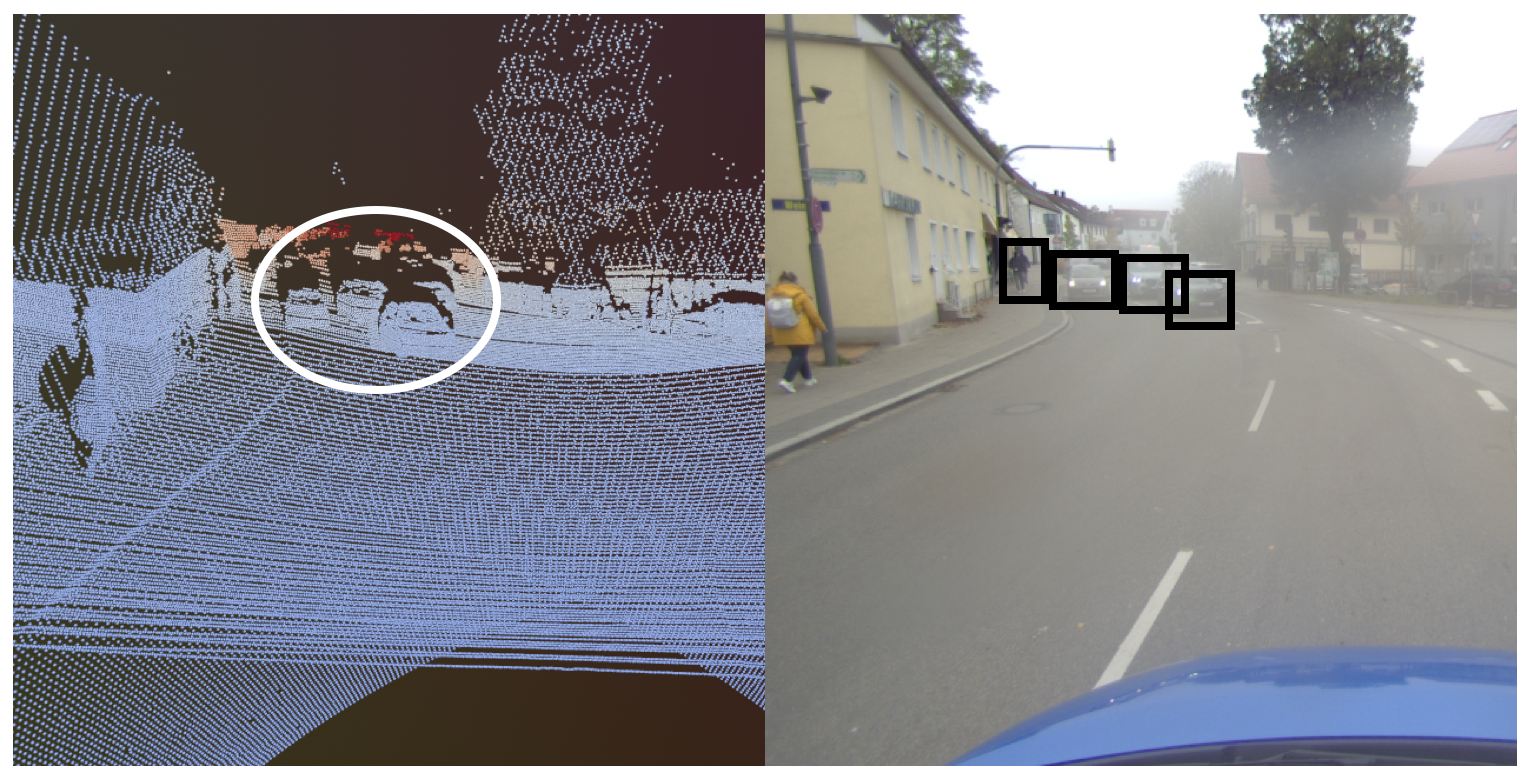}
    \caption{Typical driving scenario for an autonomous vehicle. Even for LiDAR detectors, which can measure depth directly, farther away objects are more difficult to localize, estimate and classify because of the distance and the lower number of points they reflect. Such information can be leveraged by Active Learning approaches to find the better samples to be labeled. Even human labelers require the support of camera image to label such examples, which reinforces the idea of difficulty and uncertainty.}
    \label{fig:example}
    \vspace{-0.5cm}
\end{figure}
3D object detection performs well in controlled scenarios and specific data distributions~\cite{shi2020pv, roddick2018orthographicfeaturetransformmonocular}. However, once the data distribution changes, it is necessary to record, collect and label new data, followed by retraining of the object detector. Additionally, issues like class novelty and class imbalance significantly restrict the overall performance of 3D object detection in real-world settings~\cite{geiger2012we,caesar2020nuscenes, fent2024mantruckscenesmultimodaldataset}. A straightforward approach to solve those issues involves a supervised training procedure with extensive labelled data across all scenarios. However, labelling data for every possible distribution is impractical, if not impossible. Therefore, a scalable method is essential for robust generalization of 3d object detectors across diverse environments.

Scalability can be pursued through several approaches. Among these, self-supervised and semi-supervised learning offer strategies that reduce the reliance on labelled data. However, initial human input remains necessary to provide the model with a foundational knowledge of objects, scenes, and defining characteristics. Despite this, minimizing human intervention is a key goal, making Active Learning (AL) an appealing strategy. AL focuses on identifying samples that are particularly challenging to classify or detect, enabling selective human analysis and labelling rather than annotating the entire dataset. In this way, human expertise is only required for specific, high-impact cases, delivering substantial improvements to the models in training while keeping the labelling effort manageable.

These strategies are broadly applicable across various deep learning domains; however, in the specific context of autonomous driving, they play a crucial role in 3D object detection. This application poses unique challenges, as it demands reliable detection in outdoor environments with constantly changing weather, lighting conditions, and a multitude of diverse scenarios. Additionally, while cameras are often the standard for perception tasks, LiDAR has become the preferred sensor for autonomous driving due to its advantageous 3D information capture capabilities. Although algorithms originally developed for camera-based detection, including active learning, have been adapted to LiDAR data, new challenges arise with the distinct data structure associated with this sensor. These challenges necessitate specialized approaches tailored to the unique characteristics of LiDAR-based detection~\cite{luo2023exploringactive3dobject}.

Furthermore, uncertainty estimation is a widely utilized approach by AL to identify the most optimal samples for labeling, as indicated by previous studies~\cite{hekimoglu2022, elezi2022, du2021, gal2016}. It is therefore evident that objects for which the model is less confident should provide the most value to improve its generalization capability and consequently its performance. Moreover, the results of several 3D detection architectures~\cite{sun2021, fent2024mantruckscenesmultimodaldataset} suggests that objects situated at a distance from the sensor or enclosing a smaller number of points are consistently more challenging to detect. Such practical knowledge is given in the literature already and can be heuristically implemented into the uncertainty estimation to enhance and complement it. Therefore, in this paper, we present a new active learning strategy for 3D LiDAR object detection based on both localization and classification uncertainty of augmented samples. We introduce a Gaussian mixture model (GMM) approach to calculate an inconsistency score while incorporating detection distance and class features to improve the information quality of the selected samples. 

Our contributions are summarized as below:
\begin{itemize}
    \item We propose HeAL, an explainable uncertainty estimation method derived from the real-world application of 3D detection. It uses heuristic information like object distance to enhance an augmentation inconsistency AL approach.
    \item Our sample selection strategy represents the output of the 3D detection as a Gaussian Mixture Model and therefore is agnostic to arbitratry model architectures, facilitating its implementation.
    \item We conducted several experiments with the KITTI dataset which show an improvement of 3\% mAP with respect to the current AL state-of-the-art methods (To the best of our knowledge) and the same accuracy as a fully supervised model training with 24\% of the data.
    \item We show that previous state-of-the-art approaches perform satisfactorily in a restricted data-quantity domain and propose a piece-wise approach for the AL problem, where different methods are used for different data-domains to keep a constant improvement above random sampling
\end{itemize}

%% file: 02_related.tex
\section{Related Work}
\label{sec:related}

\subsection{Active Learning}
Active learning is a strategy to minimize labeling costs by selecting the most informative samples for annotation, particularly useful in data-intensive tasks like object detection ~\cite{seung1992}. In recent years, several AL approaches have been developed, targeting the optimization of acquisition functions, uncertainty measures, and efficiency improvements in the labeling process. Classical AL methods rely on uncertainty-based sampling~\cite{du2021, caramalau2021, Gudovskiy2021, ash2020deepbatchactivelearning, gal2016}, where samples with the highest uncertainty scores are prioritized for labeling. Techniques such as BALD (Bayesian Active Learning by Disagreement)~\cite{cao2021bayesianactivelearningdisagreements} and entropy-based sampling ~\cite{wu2022} have been used to capture the most ambiguous samples for model improvement. Ensemble models, like those proposed by Beluch et al. ~\cite{beluch2018}, increase robustness by training multiple networks and aggregating predictions.

Bayesian approaches utilize Monte Carlo dropout ~\cite{houlsby2011bayesianactivelearningclassification, gal2017deepbayesianactivelearning} or variational autoencoders (VAEs) ~\cite{tran2019bayesiangenerativeactivedeep} to estimate uncertainty, which is then used as a acquisition function to select the most valuable samples. For low-dimensional problems, the uncertainty calculation is tractable, but for high-dimensional domains like image classification the uncertainty can only estimated, giving as a result possible suboptimal solutions for the samples to be labeled.

Core-set approaches~\cite{sener2018activelearningconvolutionalneural, parvaneh2022, yoo2019learninglossactivelearning} focus on sample diversity, selecting a representative subset of data points that minimizes model error on the entire dataset. Diversity-based methods aim to prevent overfitting to specific data distributions.

\subsection{Active Learning for 2D Object detection}
In image classification, uncertainty or sample-specific information is derived from the score vector. In object detection, however, additional localization and dimension data are available, helping to identify samples that are valuable for labeling. Kao et al. ~\cite{kao2019} calculate localization uncertainty by measuring the difference between the region proposal and the final object detection, which reflects the tightness of localization. Choi et al. ~\cite{Choi2021} introduce a mixture density network to generate a probabilistic distribution for both the classification and localization heads, allowing for the estimation of both aleatoric and epistemic uncertainties in the data.

\subsection{Active Learning for 3D Object detection}
The jump from 2D to 3D object detection implies in theory just depth as an extra spatial dimension and new rotation parameters. However, the weakness of monocular cameras to measure and estimate depth increases the localization uncertainty, decreasing the usefulness of the samples to be selected. To deal with this, Hekimoglu et al.~\cite{hekimoglu2022} propose a heatmap-based uncertainty estimation, predicting the position of the objects as a sum of gaussians, where the variance of the gaussians work as an uncertainty estimation. The problem with this implementation is that they use a specific architecture which outputs position as a probability distribution and therefore can not be implemented in general with any network.


In recent years, there has been a notable surge in interest in LiDAR architectures, with the aim of enhancing the localization capabilities in 3D object detection. This shift has simultaneously created a need for novel active learning approaches that can optimize data sampling and annotation efforts in this new domain. One prevalent approach is to apply active learning methodologies from classical machine learning or image classification to the domain of 3D detection. In their study, Greer et al.~\cite{greer2024} employed the entropy uncertainty measurement to identify the most optimal pointcloud samples.  However, given the distinct structure of point clouds compared to images or other forms of inputs, it is necessary to develop methods that are specifically tailored to this domain. Luo et al. ~\cite{luo2023exploringactive3dobject} prioritize label diversity and the representativeness of sampled data, particularly by leveraging point cloud density inside detected bounding boxes. The objective of this strategy is to enhance the generalization capabilities of the selected training subsets. In a subsequent study, Luo et al.~\cite{luo2023kecor}, revisited the fundamentals of information theory, selecting point clouds that maximize the minimal coding length required by a neural network to represent their latent features. Compared to them, we leverage the class, distance and point-density information directly on the scene representation. Additionally, our implementation does not require an extra proxy-network to calculate the coding rate or several forward passes to estimate the feature vector as a representativeness measure. 
Furthermore, to facilitate the implementation and evaluation of these active learning methods, Ghita et al.~\cite{ghita2024activeanno3d} and Luo et al.~\cite{luo2023exploringactive3dobject} introduce ActiveAnno3D and Active-3D-Det frameworks respectively, which are designed to enable the testing and comparison of LiDAR 3D active learning strategies.
 Lastly ,Wei et al.~\cite{wei2024} improve the performance of AL by carefully tailoring the initial set of samples used for the initial supervised step, as a form of warm start for the selection.
\subsection{Consistency-based active learning}
There are two ways to exploit inconsistency inside the AL domain: the query by committee approach, and the augmentation one. In query by committee, several models are trained with the same data and then the prediction of one input is queried. The inputs for which the models do not agree are the ones selected for labeling, as they present the maximal information gain. On the other hand, in the augmentation strategy there is only one model which is queried, but the inputs are preprocessed with light augmentations which should not change the result of the output. In the ideal case, both the original and augmented sample should output the same result, but when not, it means that the evaluated sample has a possible greater information gain and therefore should be labeled. For example, Yu et al.~\cite{yu2022} demonstrate how classification-based active learning approaches underperform when used for 2D detection tasks. They propose to augment the images and calculate a consistency metric to estimate the uncertainty of the detector. This consistency is used as a measurement of how interesting the sample is. Jeong et al.~\cite{jeong2019} applies consistency not for finding the best samples to label, but to improve the quality of pseudolabels in a semi-supervised learning approach.
For multimodal setups, with simultaneous LiDAR and Camera data, the inconsistency between the two modalities can also give hints about the most informative samples~\cite{rivera2024, yuan2023}.
More recent advancements, as presented by Elezi et al.~~\cite{elezi2022}, emphasize a framework that combines uncertainty with robustness scores for a more comprehensive acquisition function. This method considers both the uncertainty and consistency of predictions across augmentations, addressing class imbalances and distribution drift by focusing on hard-to-detect classes. This unified approach not only enhances class-agnostic performance but also incorporates auto-labeling to balance labeling costs. Nevertheless, similar to several AL approaches for classification or 2D detection, they focus on the class vector of the output to calculate the inconsistency, whereas for 3D Object detection the localization is the greatest challenge looking forward to better performance.

%% file: 03_method.tex
\section{Method}
\label{sec:method}
\subsection{Problem formulation}
Given a point cloud \( P = (x, y, z) \), where \((x, y, z)\) represents the 3D coordinates of each point, the objective of 3D object detection is to identify and localize the objects of interest as a set of 3D bounding boxes \( B = \{b_k\}_{k \in [0,N_B]} \), where \( N_B \) indicates the total number of detected bounding boxes. Each bounding box comprises the center position \((p_x, p_y, p_z)\), the dimensions \((l, w, h)\), and the orientation angle \( \theta \) with respect to the ego vehicle. The associated labels are denoted as \( Y = \{y_k\}_{k \in [0,N_B]} \in \{1, \ldots, C\} \), where \( C \) is the number of classes.

Our active learning pipeline begins with a small labeled dataset \( D_L = \{(P, B, Y)_i\}_{i \in [m]} \) and a larger pool of unlabeled data \( D_U = \{P_j\}_{j \in [n]} \), where \( m \ll n \). Additionally, we augment the original point cloud \( P \) by rotating it \( 180 \) degrees along the z-axis to generate an augmented point cloud \( P' \), resulting in the corresponding augmented bounding boxes \( B' \) and labels \( Y' \)~\cite{rivera2024, hekimoglu2022}. We choose this specific augmentation, because in the augmented version of the point cloud the represented scene is also plausible in a real scenario, compared to scaling or filtering approaches.

\begin{figure*}[tp]
    \centering
     \includegraphics[width=0.8\linewidth]{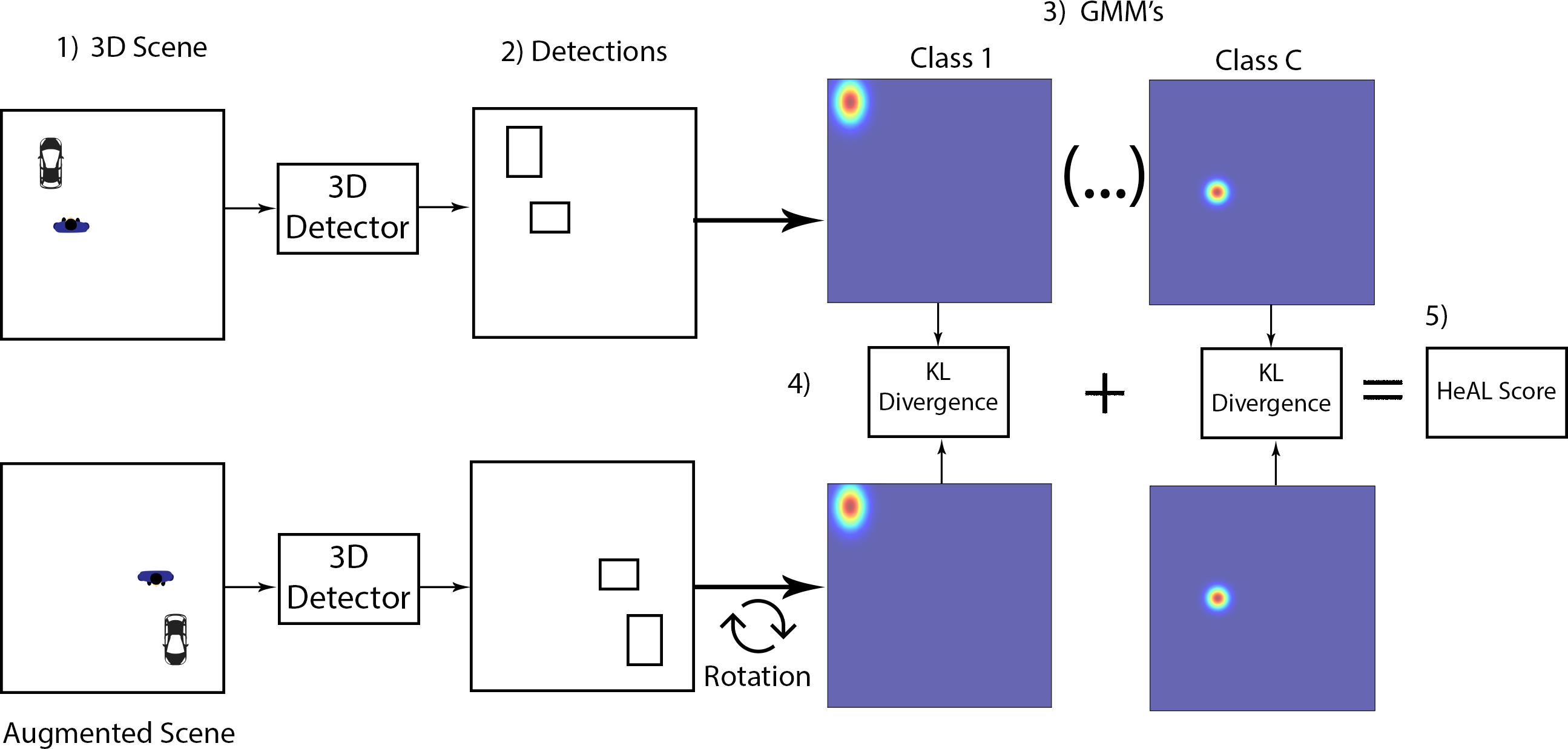}
    \caption{HeAL Score calculation pipeline, presented as BEV for clarity whereas the actual procedure is done on 3D. 1) Our input is a 3D scene represented by a point cloud, which is augmented through a 180 degree rotation around the Z axis perpendicular to the ground. 2) The 3D model from the previous AL iteration is used to detect the bounding boxes on the scene, for both the original and the augmented scene. The augmented detections are rotated again so they can be compared directly with the original detections 3) A GMM is calculated for each one of the classes in the detection, where the variance is determined by the dimensions of the object and the distance to the sensor or the point quantity of the box. 4) The inconsistency between original and augmented GMM is calculated with the KL Divergence. 5) The final HeAL score is the mean KL Divergence across all the classes. For the next AL cycle, the 100 samples with the greatest HeAL score are selected for labeling}
    \label{fig:pipeline}
\end{figure*}

During the training phase, the 3D detection model is initially trained with \( D_L \). For each active learning round \( r \in [R] \), we aim to select a subset of point clouds \( \{P_j\}_{j \in [0,N_r]} \) from \( D_U \) based on our active learning policy. We then query the labels of the 3D bounding boxes from an oracle \( \Omega: P \rightarrow B \times Y \) and combine these with the labeled dataset to form \( D_S = \{(P, B, Y)_j\}_{j \in [0,N_r]} \cup D_L \). The model is subsequently retrained with this updated dataset until the selected samples reach the final budget \( B \). We evaluate with respect to the number of samples and not to the number of boxes\cite{luo2023kecor}. In \cref{fig:pipeline} the pipeline to calculate the score is shown.

\subsection{Localization Probability Map Calculation} \label{probabilitymap}
While certain approaches calculate uncertainty by directly applying measures like Smooth-L1 Loss between predictions of original and augmented point clouds \cite{hwang2023joint}, we find it more effective to represent detected objects within a gaussian mixture model (GMM)\cite{roddick2018orthographicfeaturetransformmonocular, hekimoglu2022}. To create a robust probability map for both the original and augmented point clouds, we apply a 3D Gaussian probability density function (PDF) to the bounding box predictions generated by the detector. For each detected bounding box \( b_k \in B \) in the original point cloud \( P \) and \( b_k' \in B' \) in the augmented point cloud \( P' \), we define a probability density function that models the likelihood of object's presence within the predicted boundaries. With this approach, we do not need the network to output directly the heatmap, but we created it from the usual output of object detection models, making our approach compatible with most of 3D detection architectures.

Each bounding box \( b_k \) is represented by its center coordinates \( \mu = (p_x, p_y, p_z)\), dimensions \((l, w, h)\), and orientation angle \( \theta \). We define a 3D Gaussian distribution centered at \((p_x, p_y, p_z)\), with standard deviations proportional to the dimensions \((l, w, h)\), to capture the spatial uncertainty around the detected object center. 

The probability at the location \(\textbf{x} = \{x, y, z\}\) for a given bounding box \( b_k \) is expressed as:

 

\begin{equation} \label{eq:1}
f_k(\mathbf{x}) = \frac{1}{(2\pi)^{3/2} |\boldsymbol{\Sigma}_k|^{1/2}} \exp \left( -\frac{1}{2} 
(\mathbf{x} - \boldsymbol{\mu})^\top
\boldsymbol{\Sigma}_k^{-1}
(\mathbf{x} - \boldsymbol{\mu})
\right)
\end{equation}
\vspace{-0.5cm}
\[
\mathbf{\Sigma_k} = \begin{pmatrix}
\sigma_x^2 & 0 & 0 \\
0 & \sigma_y^2 & 0 \\
0 & 0 & \sigma_z^2 \\
\end{pmatrix}
\]
where the covariance matrix \( \mathbf{\Sigma} \) is diagonal, and \( \sigma_x^2 = l  \), \( \sigma_y^2 = w \), \( \sigma_z^2 = h \) are the standard deviations derived from the bounding box dimensions.

For each bounding box prediction \( b_k \) in \( P \) and its augmented prediction \( b_k' \) in \( P' \), we compute the 3D Gaussian PDF over a spatial grid around the bounding box center. This yields probability densities representing the likelihood distribution of an object's presence within each predicted box in both original and augmented frames. 

Then, the probability map for the sample is defined as the equally-weighted-GMM resulting of the sum of all gaussians:
\begin{equation}
    \label{eq:2}
    f(\textbf{x}) = \sum_k f_k (\textbf{x})
\end{equation}

\subsection{KL Divergence for Uncertainty Estimation}
To estimate uncertainty within our active learning framework, we utilize the Kullback-Leibler (KL) Divergence, a measure of how one probability distribution diverges from a reference distribution. In our work,  KL Divergence calculates the difference between the probability maps generated from the detections on original and augmented point clouds, capturing the spatial uncertainty around detected objects.

Given the probability map \( f(\textbf{x}) \) for a LiDAR point cloud in the original point cloud \( P \) and the corresponding probability map \( g(\textbf{x}) \) in the augmented point cloud \( P' \), we first rotate \( g(\textbf{x}) \) by 180 degrees along the z-axis to align it with the original orientation in \( P \). This alignment ensures that the probability distributions \( f(\textbf{x}) \) and \( g(\textbf{x}) \) are spatially comparable.

Once aligned, we compute the KL Divergence \( D_{KL}(f || g) \) to measure the divergence between the two probability distributions:

\begin{equation}
D_{KL}(f || g) = \sum_\textbf{\textbf{x}} f(\textbf{x}) \log \frac{f(\textbf{x})}{g(\textbf{x})} \
\end{equation}

In this equation, \( f(x) \) represents the probability map for the predictions in the original point cloud, while \( g(x) \) reflects the same in the augmented (and now aligned) point cloud. A high KL Divergence indicates a significant difference in object positioning or shape between the original and augmented frames, suggesting greater uncertainty in the model's predictions for that particular bounding box.

The KL Divergence calculated between the original and augmented probability maps serves as an uncertainty metric that informs our active learning policy to prioritize samples with higher uncertainty for labeling in subsequent training rounds. By focusing on samples with higher KL Divergence, we aim to improve the model's robustness and enhance overall detection accuracy through iterative learning.

\subsection{Classwise Probability Maps for Enhanced Uncertainty Estimation}

The creation of probability maps, as described in Section \ref{probabilitymap}, focuses on the 3D localization of the object. However, 3D object detection also requires accurately classifying the detected objects into predefined categories. To take the classification component into account for our uncertainty estimation, we introduce a novel method to incorporate classwise probability maps that capture the likelihood of each object class within the detected bounding boxes.

For each class label predicted by the 3D detection model, we generate a separate GMM for the corresponding class in both the original point cloud \( P \) and the augmented point cloud \( P' \) representing all the objects in the scene which correspond to that class. This classwise approach allows us to account for both the spatial distribution and the predicted category, enhancing the accuracy of uncertainty estimation across different object classes.

To calculate uncertainty, we compute the KL Divergence between the probability maps of corresponding classes from the original and augmented point clouds. Specifically, let \( f_c(x) \) represent the probability map for class \( c \) in the original point cloud and \( g_c(x) \) the probability map for the same class in the augmented point cloud. After aligning \( g_c(x) \) by rotating the bounding box predictions in \( P' \) by 180 degrees along the z-axis to match \( P \), we calculate the classwise KL Divergence \( D_{KL}(f_c || g_c) \) as follows:

\begin{equation}
D_{KL}(f_c || g_c) = \sum_x f_c(\textbf{x}) \log \frac{f_c(\textbf{x})}{g_c(\textbf{x})} \
\end{equation}

In this equation, \( f_c(x) \) represents the probability map for the object of class \( c \) within the original point cloud. In contrast, \( g_c(x) \) represents the same in the augmented point cloud after alignment. This process is repeated across all classes in each active learning round, yielding class-specific uncertainty scores.

To compute the overall KL Divergence \( D_{KL} \) for uncertainty estimation, we calculate the classwise KL Divergence values \( D_{KL}(f_c || g_c) \) for each class \( c \), and then take the average over all classes \( C \).

\begin{equation}
D_{KL} = \frac{1}{C} \sum_{c=1}^{C} D_{KL}(f_c || g_c)
\end{equation}
With this approach, the objects with wrong predicted classes will have a greater contribution to the general uncertainty estimation.
This provides a comprehensive uncertainty score that integrates both spatial and class information across all detected classes.

By implementing this approach of calculating classwise KL Divergence, our active learning strategy prioritizes data samples with higher uncertainty on the basis of spatial and class dimensions. This approach aims to improve the detection model’s performance on challenging classes and enhance overall accuracy in class-specific detection and localization through iterative active learning.

\subsection{Distance and Point-quantity Uncertainty Correction}

In LiDAR point clouds, distant objects are often represented by fewer points, which can affect the detector's accuracy. To address this issue, we propose two additional corrections to the initial GMMs, namely one for the distance of the bounding boxes and one for the number of points enclosed to the bounding box. Both farther-away and low-density boxes are more difficult to recognise, localise and estimate, and therefore, their uncertainty should be higher. On the one side, we propose the following correction parameter for the distance uncertainty for each box $b_k$:

\begin{equation}
U_{kd} = \frac{\sqrt{(x_k)^{2}+(y_k)^{2}}}{max_k(\sqrt{(x_k)^{2}+(y_k)^{2}})}
\end{equation}
Where $x_k$ and $y_k$ are the cartesian coordinates of $b_k$.

On the other side, we utilize the number of points within each bounding box to calculate point uncertainty. The uncertainty is determined using the following logarithmic formula:

\begin{equation}
U_{kp} = \log\left(\frac{S + \epsilon}{N_k + \epsilon}\right) + 1
\end{equation}

where \( U_k \) represents the uncertainty for bounding box \( k \), \( S \) is the total number of points across all bounding boxes, \( N_k \) is the number of points within bounding box \( k \), and \( \epsilon \) is a small constant (e.g., \( \epsilon = 1\times 10^{-9} \)) to prevent division by zero. 

This formulation ensures that bounding boxes with fewer points or larger distances are assigned greater uncertainty, effectively emphasizing distant objects that may be challenging for the detector to identify accurately. The calculated point uncertainties are integrated into the creation of Gaussian probability maps, enhancing the spatial representation of detected objects.

\begin{equation}
\begin{split}
\mathbf{\Sigma_{kd}} = U_d \cdot \mathbf{\Sigma_k} \\
\mathbf{\Sigma_{kp}} = U_p \cdot \mathbf{\Sigma_k}
\end{split}
\end{equation}

where \( \Sigma_{kp} \) and \( \Sigma_{kd} \) are the adjusted standard deviation for bounding box \( k \), \( \Sigma_k \) is the original standard deviation used in Equation \ref{eq:1}, and \( U_{kd} \) and \( U_{kp} \)  are both distance and point uncertainties.

This strategic approach ensures that even challenging classes, particularly those represented by distant objects, are adequately considered during the training process, ultimately leading to improved detection accuracy in complex environments, whose individual components are explainable.

\subsection{Experiment}

\subsubsection{Datasets}

The KITTI dataset \cite{geiger2012we} is a widely recognized benchmark for point cloud-based 3D object detection. The dataset consists of 3,712 training samples and 3,769 validation samples. It contains labels of cars, pedestrians, and cyclists. 
The nuScenes dataset \cite{caesar2020nuscenes} offers a comprehensive suite of sensory data, with precise annotations for 40,000 frames. To ensure consistency with the KITTI dataset representation, we selected a subset of 7,481 samples divided into 3,712 samples for training and 3,769 samples for validation.

\subsubsection{Evaluation Metrics}

For a fair comparison of the baselines and our proposed method, we utilize the evaluation protocol established for the KITTI dataset, following the approach of Shi et al.\cite{shi2020pv}. Specifically, we employ Average Precision (AP) for both 3D object detection, and the evaluation difficulty is categorized as EASY, MODERATE, and HARD. The related Intersection over Union (IoU) thresholds are set at 0.7 for the car class and 0.5 for both pedestrian and cyclist classes. The results are computed using 40 recall positions on the validation split. 

\subsubsection{Implementation Details}

To ensure the reproducibility of the baselines and the proposed approach,  we implemented our method using the publicly available ACTIVE-3D-DET toolbox \cite{luo2023exploringactive3dobject}. We utilize the PV-RCNN \cite{shi2020pv} backbone as the standard 3D object detection model for a fair comparison across baselines and our proposed approach. All models were trained using an NVIDIA A40 GPU. The batch sizes are set to 1 for training and 4 for evaluation on the KITTI \cite{geiger2012we} 
dataset. We adopt the Adam~\cite{adam2014} optimizer with an initial learning rate of 0.01, following a one-cycle learning rate scheduler for efficient convergence.  The code for the strategy will be available upon acceptance.

For each active learning cycle, the number of unlabeled samples to be queried was defined as $N_r=100$, training the models 30 epochs for each cycle. To evaluate 3D object detection, the 3D mAP is presented.
In each plot, the mean results for three different seeds were presented with a shaded region representing the standard deviation.


\subsection{3D Object Detector}
For our object detection pipeline, we employ the PV-RCNN model, a state-of-the-art architecture in point cloud processing for autonomous driving applications. PV-RCNN \cite{shi2020pv} combines voxel-based and point-based feature extraction methods to achieve high accuracy in 3D object detection. The model’s architecture utilizes both 3D voxel CNNs and PointNet~\cite{qi2014} to effectively capture and encode fine-grained spatial features, which enables robust and accurate predictions of bounding boxes around objects within point cloud data.

Our pipeline begins by feeding raw point cloud data into PV-RCNN to obtain initial predictions. These predictions consist of bounding boxes, object classes, and associated confidence scores. In the following steps, these predictions serve as the basis for uncertainty estimation and sample selection in an active learning framework.

%% file: 04_results.tex
\section{Results}
\label{sec:results}

For the comparison with the literature, we compared several methods: First, Random sampling as baseline; for domain agnostic AL, Entropy\cite{wang2014} and Montecarlo\cite{feng2019}; Finally, for LiDAR 3D-Detection methods we compared with the current State-of-the-art, namely CRB\cite{luo2023exploringactive3dobject} and KECOR \cite{luo2023kecor}.

\begin{figure}[ht]
    \centering
    \includegraphics[width=\linewidth]{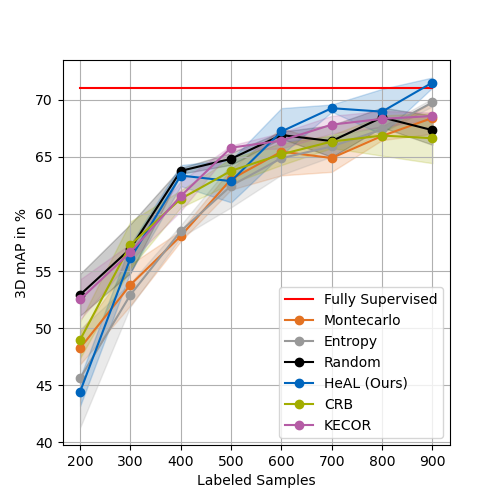}
    \caption{mAP in \% with respect to the number of labeled samples for selected AL strategies on KITTI.}
    \label{fig:mmap}
    \vspace{-0.5cm}
\end{figure}
Figure \ref{fig:mmap} shows the results across eight active learning cycles. In the low-data regime, only KECOR has the same performance as the Random baseline. However, as more samples are incorporated, HeAL demonstrates increasingly robust results, outperforming both KECOR, CRB and the Random baseline by 3\%, 4.9\% and 4.2\% after the eighth active learning cycle, respectively. Notably, in the final iteration, even the simple entropy-based selection strategy surpasses KECOR in performance. Discriminating according to difficulty level confirms the trend, where for the 3 categories a similar trend as in the combined mAP is observed.

\begin{figure}[ht]
    \centering
        \includegraphics[width=\linewidth]{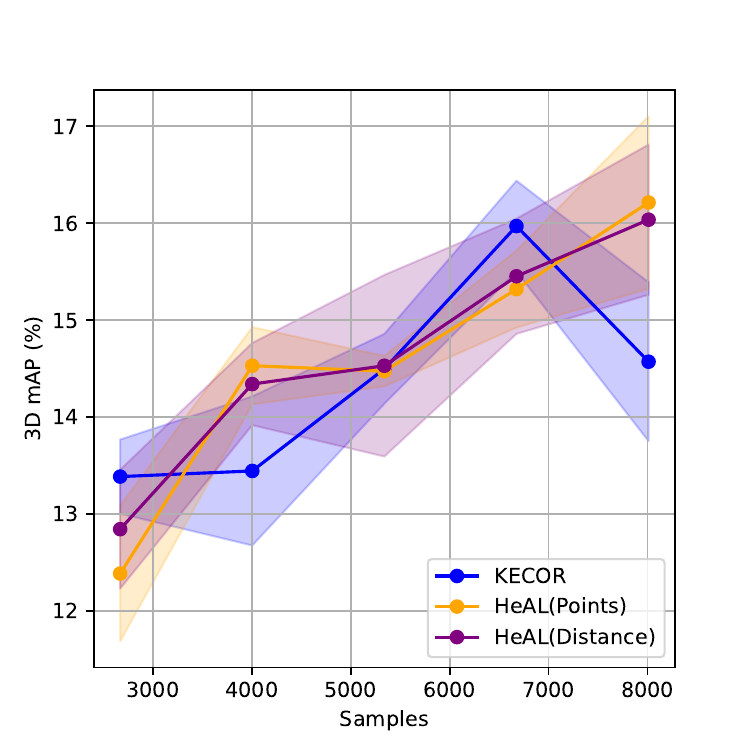} 
        \caption{mAP in \% with respect to the number of labeled samples for selected AL strategies on Nuscenes.}
        \label{fig:mmap NuScenes}
        \vspace{-0.5cm}
\end{figure}

We provide an evaluation of both Heal and the state-of-the-art KECOR method using the NuScenes dataset in \Cref{fig:mmap NuScenes}, where our approach consistently outperforms KECOR by 2\% across three different seeds. 
Furthermore, its performance across iterations demonstrates a clear upward trend, in contrast to KECOR, which exhibits more unpredictable behavior. This pattern aligns with the evaluation of KECOR and CRB on the Waymo dataset~\cite{Sun_2020_CVPR}, where their trends are similarly harder to discern. This variability can be attributed to the overall complexity of datasets like NuScenes and Waymo, which pose significant challenges for detectors. 

\begin{figure*}[ht]
    \centering
    \includegraphics[width=0.9\linewidth]{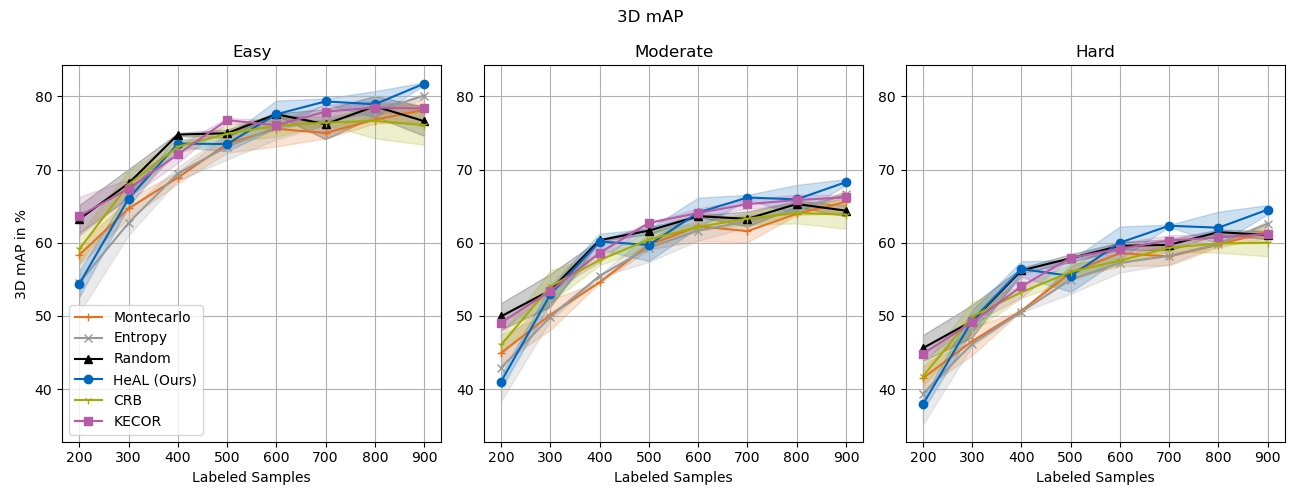}
    \caption{mAP in KITTI for each difficulty level}
    \label{fig:map}
    \vspace{-0.5cm}
\end{figure*}

\begin{figure*}[ht]
    \centering
    \includegraphics[width=0.9\linewidth]{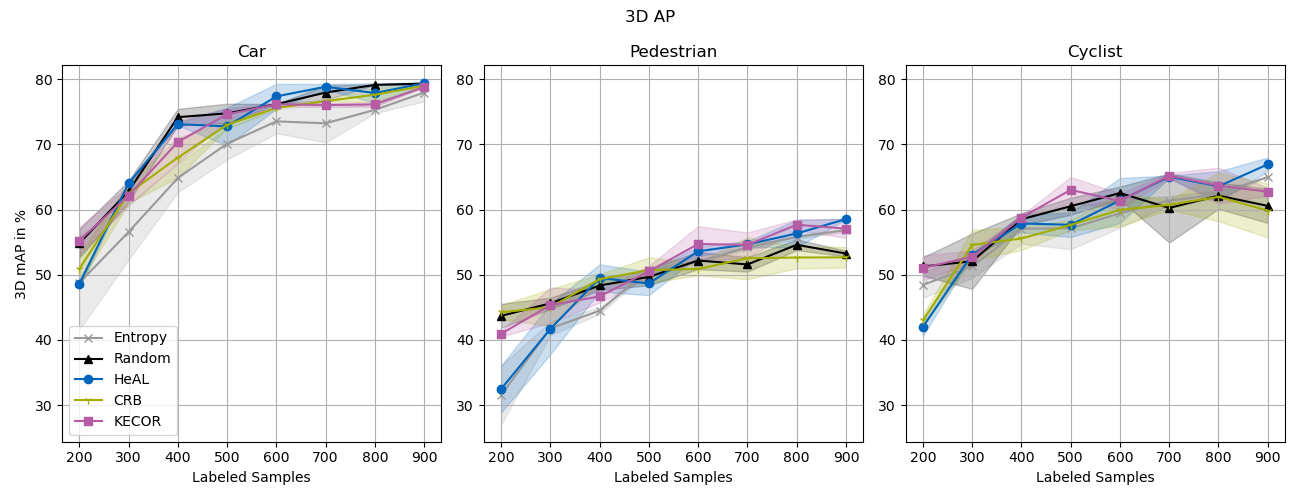}
    \caption{AP in KITTI for each class}
    \label{fig:classwise}
    \vspace{-0.5cm}
\end{figure*}
In \cref{fig:map}, HeAL outperforms the concurrent method for all difficulty levels. Similarly, separating the results by class (\cref{fig:classwise}) shows an interesting behavior for HeAL: For the class Car, which is the most representative class in KITTI, there is almost no improvement with respect to the random baseline; however, for both pedestrian and cyclist there is a clear improvement on the AP. The sample selection method improve the performance of unrepresented classes, mostly because for them is easier to find extra information to improve the detection, rather than from the Car class, where there are already too many samples to start with.

More generally, in the higher data regime, both CRB and KECOR demonstrate minimal to no improvement, whereas HeAL and even entropy (in the final iteration) exhibits superior performance. It is evident that KECOR exhibit its greatest strength during the initial iterations, with a reduced number of boxes, and present the most significant positive difference in performance. However, the trend indicates that for the higher-data regime, their performance reaches a plateau, indicating that no further improvement can be extracted from the information-theoretic insights they propose. We argue that, given the current state of the art in this field, there is no optimal AL strategy for the entire data range. Rather, it may be more beneficial to employ a combination of strategies for different data domains, particularly when working with a fixed budget. Specifically, KECOR can be employed as a preliminary approach, followed by a transition to an alternative strategy if the budget permits and their performance does not improve. This alternative could be HeAL which can enhance the model's performance through the whole range. This complements the goal of AL, which is to obtain the highest possible metrics with the smallest number of samples. Even if the labeling budget is not extremely small, the goal is still to find the best samples to label in the whole data space: HeAL is then a good complementary strategy.

\subsection{Ablation studies}

\begin{table}[h!]
    \centering
    \vspace{-0.2cm}
    \renewcommand{\arraystretch}{1.2} 
    \setlength{\tabcolsep}{4pt} 
    \caption{Ablation mAP results for 700 samples on KITTI}
    \resizebox{0.5\textwidth}{!}{%
    \begin{tabular}{c c c c|c c c c }
        \hline
        GMM & Class & Distance & Point & Easy & Moderate & Hard & Total \\
        \hline
        \checkmark   &       &          &       &75.9&   63.3 &   59.5   &66.3 \\
        \checkmark   & \checkmark &     &       &   77.2    &   63.8     &  59.9    & 67.0 \\
        \checkmark   &  &  \checkmark   &       &   74.5    &   61.1     &  57.1   & 64.3 \\
        \checkmark   &       &          & \checkmark &75.4 & 61.87 & 57.9     &65.0  \\
        \checkmark   & \checkmark &     & \checkmark & 76.5 & 62.8    &   58.9   & 66.1 \\
        \checkmark   & \checkmark & \checkmark &     & \textbf{79.3} &    \textbf{66.2}  &   \textbf{62.3}   & \textbf{69.3} \\
    \end{tabular}}
    \label{tab:ablation}
\end{table}
Finally, we investigate separately the different modules which compose our method, namely localization uncertainty, classification correction and difficulty correction. In \cref{tab:ablation} we show which separated modules contribute the most to the final performance. It can be seen, that the point uncertainty presents the lowest performance coupled directly with the localization uncertainty, being even lower than the vainilla localization uncertainty. In contrast, the class information improves the pure heatmap in 1.3\%. The best performance is shown by the Localization+Class+Distance uncertainty approach, which achieves an improvement of 2.3\% with respect to the second best, the Localization+Class approach.
The separation of Class into multiple GMM appears to be the defining factor to improve the performance. This could be because this separation penalizes more strictly a missclassification of the object than differences in the classification score vector of a regular object detector.

Lastly, even though both the distance and number of points are correlated, it is clear that the number of points damages the overall efficiency of the sampling strategy. A possible explanation, is that occlusions cause that some objects, even when they enclose few points, are still close enough to the sensor, and are therefore easier to localize, classify and detect than farther away boxes, which would make them less suitable to be labeled, given their low amount of valuable information for the training.

%% file: 10_conclusion.tex
\section{Conclusion}
\label{sec:conclusion}

In this paper we present HeAL, a novel and explainable AL sampling method for 3D Object detection. Our method is derived from heuristic observations regarding the difficulty of detecting objects located at a distance. This information is then used to adjust the uncertainty of a GMM representing the detection heatmap from a scene. Subsequently, this heatmap is compared with an augmented version of it, with the aim of selecting the samples exhibiting the greatest inconsistencies. Our approach achieves results competitive to the state-of-the-art  for the data domain where previous works have reached a plateau. We therefore propose a multi-approach strategy, in which several AL strategies are employed sequentially to achieve the largest mAP improvement across the entire data range. 

%% file: 12_appendix.tex
\section{Appendix Section}
\label{sec:appendix_section}
Supplementary material goes here.